\documentclass{article}

\usepackage[main, final]{neurips_2026}

\usepackage[utf8]{inputenc} 
\usepackage[T1]{fontenc}    
\usepackage{hyperref}       
\usepackage{url}            
\usepackage{booktabs}       
\usepackage{amsfonts}       
\usepackage{nicefrac}       
\usepackage{microtype}      
\usepackage{xcolor}         
\usepackage{graphicx} 
\usepackage{amsmath}
\usepackage{multirow}
\usepackage{placeins}
\usepackage[table]{xcolor}
\usepackage{wrapfig}
\usepackage{caption}
\usepackage{makecell}
\usepackage{fontawesome5}   

\title{P$^3$-VLM: A Point-based Alternative for Grounded \\ 3D Vision-Language Models}

\author{%
  Anna-Maria Halacheva,
  Jan-Nico Zaech, Sombit Dey, Luc Van Gool, Danda Pani Paudel\\
  INSAIT, Sofia University “St. Kliment Ohridski” \\
  \\
  \faGlobe\ \href{https://p3vlm.insait.ai/}{\texttt{https://p3vlm.insait.ai/}} \\
}

\begin{document}


\maketitle
\vspace{-35pt}
{
\begin{center}
    \centering
    \includegraphics[width=\linewidth, trim={0cm 0cm 0cm 0cm}, clip]{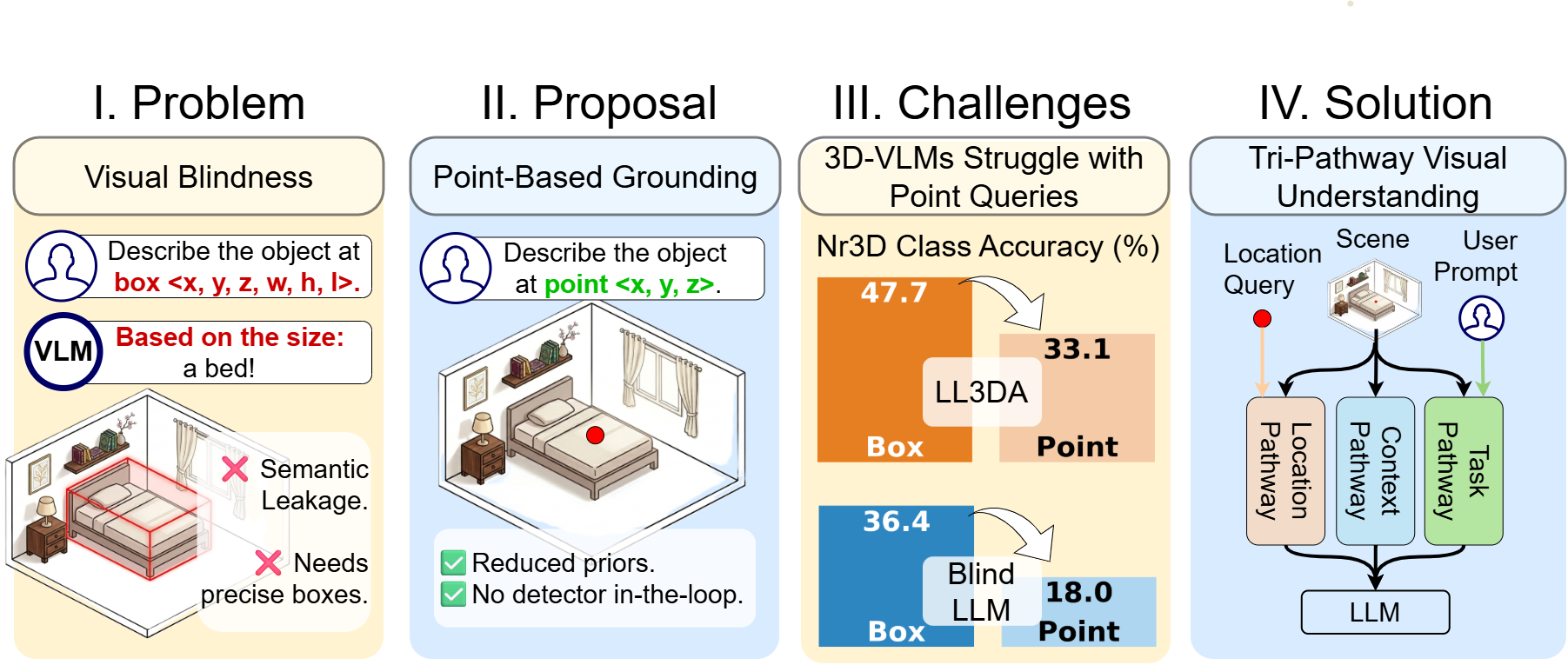}
    \captionof{figure}{We address bounding box induced semantic leakage and thus the visual blindness in {3D VLMs} using a point-based grounding alternative.
    This is challenging because both 3D VLMs and blind LLMs struggle with point queries. To overcome this, we develop a novel tri-pathway 3D VLM. }
    \label{fig:teaser}
\end{center}
}
\begin{abstract}
   Examining existing benchmarks for grounding in 3D VLMs reveals a pervasive ``semantic leakage" in the evaluation and prompting protocols:
   a heavy reliance on bounding boxes as spatial prompts or representation primitives. 
   Because bounding boxes are highly informative of object categories, they introduce a shortcut for the models to neglect visual information during 3D VLM training. 
   We demonstrate this with a ``blind" LLM that, prompted only with the box coordinates, can perform surprisingly well.
   To further analyze this shortcut, we propose \emph{point-based grounding prompting} -- querying models with a single 3D point instead of a box. Under this protocol, the accuracy of the blind LLM drops from $\textbf{36.4\%}$ to $\textbf{18.0\%}$ on ScanNet's Nr3D, and similarly degrades for 3D VLMs (e.g. LL3DA).
   
   Building on this insight, we introduce \textbf{P$^3$-VLM}, a novel tri-pathway 3D VLM that supports point-based prompts and achieves $\textbf{52.1\%}$ state-of-the-art accuracy on location-grounded captioning under this setting. To achieve this, we redesign the architecture while keeping it detector-free, scene-centric, and suitable for 3D GS inputs. P$^3$-VLM utilizes three complementary pathways (location-, task-, and context-aware), enabling strong reasoning about scenes without any bounding box dependencies. The key benefit of point-based learning is a consistent improvement across settings: P$^3$-VLM achieves state-of-the-art performance under point-based grounding while also improving robustness, OOD generalization, and performance on ungrounded scene-centric 3D VQA benchmarks, reflecting its reduced reliance on bounding-box shortcuts and resulting stronger visual-spatial reasoning. 
\end{abstract}

\section{Introduction}
3D Vision-Language Models (VLMs) hold the promise of generalist scene understanding: given a 3D representation of an environment, such a model should be able to answer both grounded queries -- those anchored to a specific location or object in the scene -- and ungrounded queries about the scene at large \cite{chen2024ll3da,huang2023embodied,chen2024ll3da,3dllm,huang2024chat}. Significant progress has been made in this direction, yet a fundamental question remains open: to what extent do current 3D VLMs actually observe the scene, rather than exploit contextual shortcuts?

We investigate this question through grounded object understanding: identifying the correct object given a 3D spatial prompt. This requires jointly reasoning about geometry and semantics, including localizing the queried region, interpreting its visual content, and producing a semantically accurate response \cite{chen2021scan2cap,chen2020scanrefer,halacheva2025gaussianvlm,mei2025PerLA}.
We find that existing object- and scene-centric paradigms share a common issue of “semantic leakage” in evaluation protocols, arising from the widespread use of bounding boxes as either prompts or representation primitives. Because bounding boxes carry strong categorical cues through their size, shape, and aspect ratio, they enable shortcut solutions (e.g., a tall narrow box often corresponds to a door; beds have standardized measurements) that reduce the need for true visual-spatial reasoning. They also implicitly assume accurate object localization, which is usually unrealistic in real-world 3D applications.

The box dependency appears across both object-centric VLMs \cite{huang2023embodied,huang2024chat,zhu2025unifying}, which inherit box-based object representations from detectors, and scene-centric models \cite{halacheva2025gaussianvlm,chen2024ll3da}, which still rely on bounding boxes as grounding prompts at inference time. As a result, models can exploit categorical priors encoded in box geometry rather than visual evidence. We demonstrate this effect by showing that a blind LLM trained only on bounding box coordinates, without any visual input, achieves $36.4\%$ accuracy on Nr3D  as shown  in {Fig.~\ref{fig:teaser}. We further show that the overreliance on bounding boxes makes the existing 3D VLMs fragile, leading to significant degradation in scenarios with atypical object sizes, noisy reconstructions, or poorly aligned bounding boxes in Fig.~\ref{fig:teaser2} (robustness analysis).

These findings motivate both a new evaluation protocol and a new model design. We propose \emph{point-based grounding prompting}, where queries are specified using only a single 3D point on the target object. Unlike bounding boxes, point prompts do not reveal object shape or extent, substantially reducing category-specific geometric cues and incentivizing models to rely on genuine visual-spatial understanding. Under this stricter protocol, the shortcut enabled by bounding boxes largely disappears: the accuracy of the blind LLM drops sharply from $36.4\%$ to $ 18.0\%$ on Nr3D. Existing 3D VLMs, including scene-centric models such as \cite{halacheva2025gaussianvlm,chen2024ll3da}, also degrade substantially, further exposing their overreliance on bounding-box cues (e.g. LL3DA drops from $47.7\%$ to $ 33.1\%$, see Fig.~\ref{fig:teaser}).

To address this new challenging setting, where existing scene-centric 3D VLMs struggle, we introduce \textbf{P$^3$-VLM}, a detector-free, scene-centric 3D VLM built on 3D Gaussian scene representations (3D GS) and designed specifically for point-based grounding. P$^3$-VLM utilizes a tri-pathway architecture consisting of: (1) a \emph{location-aware} pathway that encodes compact positional cues from the query point, (2) a \emph{task-aware} pathway that sparsifies and refines query-relevant volumetric features, and (3) a \emph{context-aware} pathway that summarizes the global scene for grounded reasoning. By anchoring visual features directly to scene voxels rather than detector-derived geometry, P$^3$-VLM enables spatially grounded reasoning under the proposed point-based alternative.

Under point-based grounding, P$^3$-VLM achieves $52.4\%$ class accuracy on Nr3D, substantially outperforming both the blind LLM baseline ($18.0\%$) and previous 3D VLMs adapted to this setting ($33.1\%$ and $29.3\%$ for LL3DA and GaussianVLM, respectively; see Tab.~\ref{tab:point_vs_box}). This large margin demonstrates that its visual encoder is genuinely load-bearing. Moreover, by avoiding bounding-box shortcuts, P$^3$-VLM is not only more robust but also exhibits stronger scene-level OOD generalization (see Fig.~\ref{fig:teaser2}), while maintaining consistent gains on ungrounded scene understanding benchmarks such as ScanQA \cite{scanqa_22_cvpr} and SQA3D \cite{ma2022sqa3d}.
Overall, this work makes the following contributions:
\begin{itemize}
    \item We identify bounding box leakage as a pervasive shortcut in current 3D VLMs, hurting visual-spatial understanding, and introduce point-based grounding as a stricter protocol.
    
    \item We present P$^3$-VLM, a detector-free, scene-centric 3D VLM operating on 3D Gaussian representations via three complementary pathways (location, task, and context aware).
    
    \item We show that P$^3$-VLM achieves SOTA performance under point-based grounding while improving robustness, OOD generalization, and performance on VQA benchmarks.
\end{itemize}

\begin{figure}[ht]
    \centering
    \includegraphics[width=\linewidth]{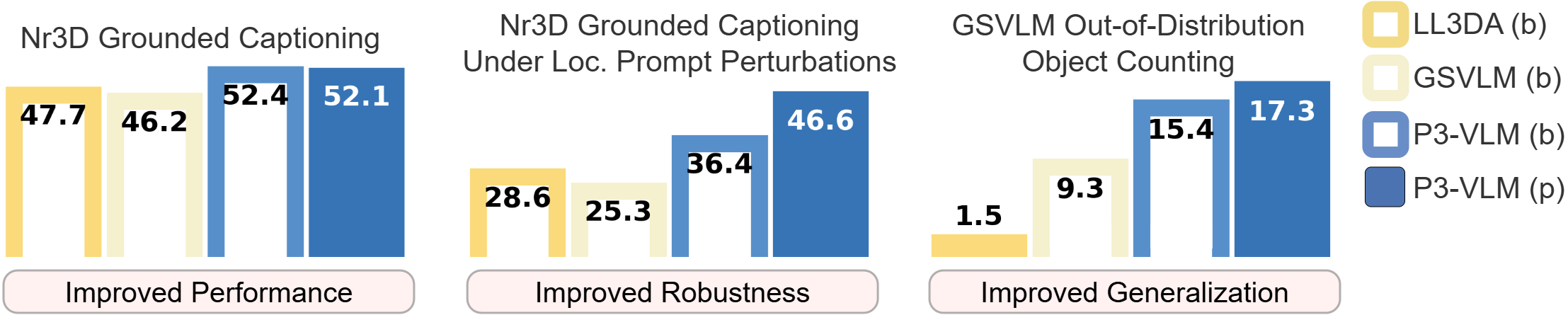}
    \caption{\textbf{Performance, robustness, and generalization.} Our P$^3$-VLM consistently outperforms VLM baselines LL3DA~\cite{chen2024ll3da} and GaussianVLM~\cite{halacheva2025gaussianvlm} (GSVLM). Notably, P$^3$-VLM on points (p) offers better robustness and generalization than box-based methods (b), despite less spatial information. }
    \label{fig:teaser2}
    \vspace{-10pt}
\end{figure}
\section{Related Work}

\subsection{3D-Grounded Scene Understanding}
3D-grounded scene understanding refers to tasks involving visual grounding in 3D space, which can manifest on either the input or the output side of the model \cite{chen2020scanrefer, achlioptas2020referit_3d,halacheva2025gaussianvlm,SpatialLM,li2025scenesplat}. 
On the output side, the model receives a reference to an object or category and must predict a spatial specification for the matching object(s) -- a 3D bounding box or segmentation mask \cite{SpatialLM,lazarow2025cubify,li2025scenesplat,ma2025scenesplat++,avetisyan2024scenescript,roomformer,kolodiazhnyi2025unidet3d,shen2024vdetr}. Specialist models for this setting \cite{lazarow2025cubify,li2025scenesplat}, including those built on language backbones \cite{SpatialLM,roomformer,avetisyan2024scenescript}, achieve strong performance by combining task-specific pretraining, structured output modeling, and dedicated detection heads. However, because of this specialization, the state-of-the-art methods sacrifice their free-form answer and thus generalist capabilities.
Generalist models with multi-task open-language capabilities \cite{chen2024ll3da,huang2024chat,huang2023embodied} have been unable to match specialist performance on output-side grounding, lacking, for instance, the structured output modeling and full-scale task-specific supervision.  
Furthermore, existing free-form benchmarks for diverse scene understanding are largely restricted to ScanNet-scale datasets~\cite{scanqa_22_cvpr,ma2022sqa3d,3dllm,huang2023embodied,jia2024sceneverse,chen2020scanrefer,achlioptas2020referit_3d,halacheva2025gaussianvlm}, hindering the training of generalist models for output-side grounding at the scale necessary to achieve state-of-the-art performance \cite{SpatialLM}.

Input-side grounded understanding entails providing the grounding on the input side through a 3D location query, with model output -- a caption of the referenced object \cite{halacheva2025gaussianvlm,chen2024ll3da,mei2025PerLA,zhu2025unifying,huang2023embodied,zhi2024lscenellm}. This setting is more compatible with open-form answer models. Some free-form VLMs also report bounding box outputs in this setting, yet those are obtained via independent, frozen detectors \cite{chen2023end,schult2023mask3d} entirely separate from their core architecture \cite{chen2024ll3da,huang2024chat,huang2023embodied}, making the box prediction not a product of the model's learned visual understanding. Exploring free-form 3D generalist VLMs, we operate in the input-side captioning setting \cite{halacheva2025gaussianvlm,mei2025PerLA,zhi2024lscenellm}. The captioning quality directly measures the VLM's visual understanding, not focusing on box predictions, obtainable through an external module.

\subsection{3D VLMs for Multi-Task Scene Reasoning}
VLMs for 3D scenes aim to support a broad range of tasks, including object captioning \cite{chen2020scanrefer,achlioptas2020referit_3d}, question answering \cite{scanqa_22_cvpr,3dllm}, spatial reasoning \cite{ma2022sqa3d,zhang2024spartun3d} and beyond \cite{halacheva2025gaussianvlm,3dllm}, through a single unified model \cite{halacheva2025gaussianvlm,chen2024ll3da,Zheng_2025_CVPR,qi2026gptscene}.
The dominant paradigm in this setting has been 
object-centric: detect a set of object proposals, encode their point 
cloud features to separate tokens, and align them with language \cite{zhang2024spartun3d,huang2023embodied,huang2024chat,li2025m3dbench,zhu20233d,wang2023chat3d,Zemskova_2025_ICCV,chen2024grounded,Wang_2024_CVPR,linghu2024msr3d,chen2024grounded}. Prior research suggests that this approach is fundamentally limited as scene-level understanding is bounded by detector recall \cite{halacheva2025gaussianvlm}, and is emergent rather than by design~\cite{halacheva2025gaussianvlm,zhi2024lscenellm,Peng2023OpenScene,delitzas2024scenefun3d}.
Thus, a parallel line of work moves away from detector dependencies, spanning video VLMs for 3D understanding \cite{fan2025vlm3rvisionlanguagemodelsaugmented,Zheng_2025_CVPR,qi2026gptscene,hu2025g2vlmgeometrygroundedvision,chen2025vid,Yang_2025_CVPR,Wang_2025_ICCV}, and native 3D models. 
The latter processes 3D scenes either (1) in a region-based manner requiring incorporation of advanced methods for instilling cross-regional awareness (e.g., GCNs \cite{mei2025PerLA}) or (2) via scene-centric strategies that encode the environment at once as a whole  \cite{halacheva2025gaussianvlm,chen2024ll3da,zhi2024lscenellm}.
GaussianVLM~\cite{halacheva2025gaussianvlm} pioneered the use of 3D GS for scene-centric understanding and acts as a baseline for this work. However, it suffers from reduced robustness and generalization due to reliance on semantic shortcuts (Sec. \ref{sec:pointVsBox},\ref{sec:exp_ood}, \ref{supp_sec:vs_gaussianvlm}).    
%

\section{From Bounding Box Semantic Leakage to Point Grounding}
A foundational assumption underlying 3D VLM evaluation is that grounded query performance reflects genuine visual-spatial reasoning -- an assumption which we investigate in this section. When models are queried about a specific object via a bounding box prompt -- the universal convention across 3D VLMs -- the box itself encodes strong categorical information: its dimensions, aspect ratio, 
and volume are highly predictive of object class. A tall, narrow box standing up from floor level implies a door; a box with measurements of a queen-sized bed -- a bed, etc. We term this phenomenon \emph{bounding box semantic leakage} and demonstrate that it acts as a dominant signal that can bypass visual perception in 3D VLMs when it comes to grounding tasks.

\paragraph{3D-Grounded Object Captioning.}
To investigate this phenomenon, we focus on the task of 3D-Grounded Object Captioning, which requires a model to generate a natural language description for a specific object given a spatial query. We evaluate various 3D VLMs on two ScanNet-based benchmarks: Nr3D \cite{achlioptas2020referit_3d} and ScanRefer \cite{chen2020scanrefer}. Firstly, we evaluate models under the standard prompting protocol, \emph{bounding-box prompting}, specifying the query location as a box center and extent.

\subsection{Bounding Box Semantic Leakage}
\begin{wraptable}{r}{0.48\textwidth}
    \vspace{-0.5cm}
    \caption{Object-centric VLMs do not utilize their 3D encoder. A blind LLM operating on bounding box coordinates alone, and an LLM with only box coordinate embeddings, perform close to the full OD-VLM on Nr3D.}
    \centering
    \small
    \begin{tabular}{lc}
    \toprule
    Model & Class Acc. (\%) \\
    \midrule
    Blind LLM (box coords only) & 36.4 \\
    Box encoder + LLM           & 44.2 \\
    Full OD-VLM (LEO)           & 45.1 \\
    \bottomrule
    \end{tabular}
    \label{tab:od_exp}
    \vspace{-0.5cm}
\end{wraptable}

We aim to isolate the contribution of geometric priors from other key aspects of 3D grounding such as semantics and context awareness. 
For this, we conduct two controlled experiments on Nr3D.
Specifically, we examine the effect of the strong reliance on bounding boxes within the textual and/or the visual component of the prompt, highlighting that the leakage can occur via both information channels and cause over-reliance on this geometric prior rather than utilization of rich visual features.

\paragraph{Leakage via Textual Prompting.} First, we examine the effect of bounding boxes as grounding prompts within the textual LLM task prompt. We fine-tune a "blind" LLM (Qwen3-VL 4B Instruct LLM \cite{bai2025qwen3}) via LoRA to predict object classes from box center, yaw and extent alone, with no visual encoder of any kind. The fine-tuned blind LLM achieves 36\% class accuracy -- a significant result that establishes box geometry as a strong categorical signal independent of any visual observation.

\paragraph{Leakage via Architectural Priors.}
Leakage can also occur through the visual channel, particularly in Object-Detector VLMs (OD-VLMs). These models typically use a decoupled architecture: a detector segments the scene, and a 3D encoder transforms each instance's point cloud into tokens for the LLM. We investigate whether the 3D encoder contributes meaningful semantic understanding beyond what the detector's box metadata already provides.

We ask whether the visual encoder in a full object-centric VLM is strongly utilized on top of this prior. For this, we evaluate an OD-VLM (LEO \cite{huang2023embodied}) against a modified version where the visual features are simply encodings of the bounding boxes (center and extent) of the objects within the scene, derived with a lightweight 2-layer MLP. 
We observe very close performance of the two models (Tab.~\ref{tab:od_exp}). These results suggest that, under bounding-box-based visual prompting, object class predictions are once again driven primarily by geometric priors rather than visual features.

\subsection{Point-Based Grounding} 
\label{sec:pointVsBox}
Our findings indicate that the prevalent use of bounding boxes as spatial prompts may provide geometric priors that simplify the grounding task. To isolate visual perception from these structural cues, we propose \emph{point-based grounding prompting}: a protocol where location queries are specified as a single 3D coordinate $(x, y, z)$. By removing information about the target’s extent or aspect ratio, it serves as a more faithful benchmark for a model’s ability to extract semantics from visual features.

The effect of removing box geometry is immediate. We fine-tune two variants of the blind LLM: one trained on bounding box prompts and evaluated accordingly, and one trained and evaluated on point-based prompts (Supp. Sec. \ref{supp_sec:exp_tunePoint}). Class accuracy drops from 36.4\% to 18.0\% (Tab.~\ref{tab:point_vs_box}) -- a 50\% relative reduction achieved by changing only the prompt format, with no modification to the model architecture. This confirms the prior inherent in the bounding-box grounding prompts, and its reduced effect in point prompts.

We then ask whether existing scene-centric VLMs can close this gap when adapted to the new prompt. For each model, we fine-tune on point-based grounding and re-evaluate under the stricter protocol (training details in the Supplementary \ref{supp_sec:exp_tunePoint}). The results in Tab.~\ref{tab:point_vs_box} show that fine-tuning does recover some performance, yet a substantial gap relative to the box-based baseline persists across all models: on Nr3D LL3DA drops from 47.7\% to 33.1\%, and GaussianVLM from 46.2\% to 29.3\%. The fact that explicit training on point prompts does not close this gap suggests that the limitation is not one of prompt formatting, but of visual capacity -- these models have not developed the dense visual representations that point-based grounding demands.

\begin{wraptable}{r}{0.50\textwidth}
    \vspace{-0.5cm}
    \centering
    \footnotesize
    \caption{\textbf{Object captioning performance on Nr3D and ScanRefer with different prompting strategies.} The blind LLM's sharp drop confirms that bounding box geometry was the dominant signal. Also the 3D VLM baselines (LL3DA, GaussianVLM) trained on boxes drop considerably even after fine-tuning on point prompts.}
    \begin{tabular}{@{\hskip 0mm}l l | c c c c}
    \toprule
      Dataset & Model & \multicolumn{2}{c}{Box} & \multicolumn{2}{c}{Point}\\
      \cmidrule{3-4} \cmidrule{5-6}
      & & Acc & Sim & Acc & Sim\\
      \midrule
      \multirow{3}{*}{Nr3D}
      & Blind LLM    & 36.4 & -    & 18.0 & -    \\
      & LL3DA        & 47.7 & 48.1 & 33.1 & 39.5 \\
      & GaussianVLM  & 46.2 & 48.2 & 29.3 & 35.5 \\
      \midrule
      \multirow{3}{*}{\makecell{Scan\\Refer}}
      & Blind LLM    & 38.1  & -    & 18.4 & -    \\
      & LL3DA        & 48.02 & 55.9 & 29.2 & 34.9 \\
      & GaussianVLM  & 50.4  & 59.1 & 33.2 & 41.4 \\
      \bottomrule
    \end{tabular}
    \label{tab:point_vs_box}
    \vspace{-0.5cm}
\end{wraptable}

\section{P$^3$-VLM}
\label{sec:method}
\begin{figure}
    \centering
    \includegraphics[width=\linewidth]{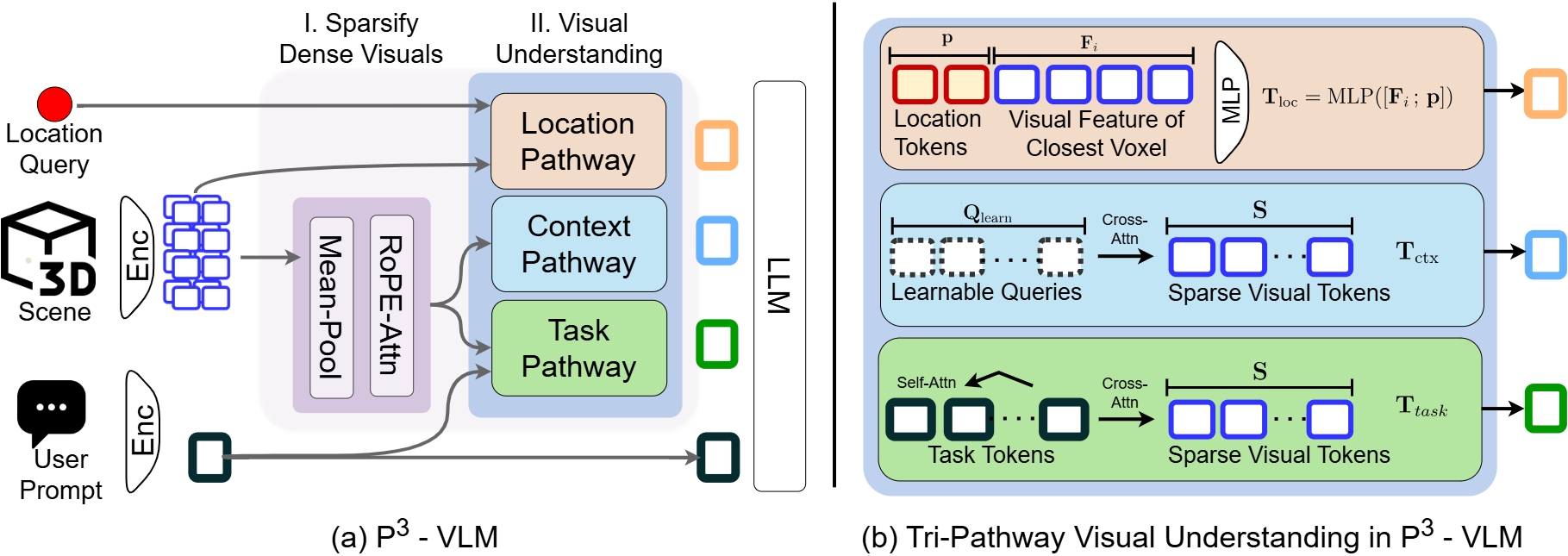}
    \caption{\textbf{P$^3$-VLM architecture overview.} Three complementary pathways process the voxelized 3D scene: the location-aware pathway extracts a single token at the queried point; the task-aware pathway produces prompt-guided scene tokens by first refining the task prompt tokens and subsequently utilizing them for visual refinement of the sparsified features;
    the context-aware pathway distills 8 global scene summary tokens via unconditional cross-attention. Notably, the context and task pathways operate on the sparsified features for efficiency gains.}
    \label{fig:method}
    \vspace{-10pt}
\end{figure}

P$^3$-VLM is a scene-centric 3D VLM, that takes as input 3D Gaussian Splats (GS) and a natural language query and produces a free-form language response relying on point-based spatial prompting.
An overview of the architecture is shown in Fig.~\ref{fig:method}. 
The scene is first voxelized: a sparse 3D GS encoder processes the raw Gaussian splats into a set of voxel features $F\in \mathbb{R}^{N\times D}$ with associated 3D coordinates $C\in \mathbb{R}^{N\times 3}$, where each feature is anchored to a specific spatial location. These voxel features then feed three complementary pathways: location-aware, task-aware, and context-aware, which together produce the visual token sequence as input to the LLM. To the visual tokens, we append the user language task tokens $L$.

\subsection{Tri-Pathway Visual Processing}
The 3D visual encoder produces rich, yet dense, visual features. Even for our downsampled scans, this can produce more than 10k tokens, which are computationally heavy to be processed as a sequence in the attention mechanism.
To alleviate this, we sparsify the dense visual features by aggregation and refinement, which additionally increases their spatial awareness, to make them consumable by the context and task pathways. 
Notably, the location pathway operates on raw encoder features as it focuses exclusively on locality rather than global awareness. 
Detailed implementation specifications and hyperparameter choices are provided in Supp. Sec. \ref{supp_sec:architecture}.

\paragraph{Sparsification.}
The raw voxel features are spatially downsampled to at most 2048 tokens per scene via average pooling on a uniform 3D grid, making subsequent self-attention tractable.
The downsampled features are then spatially enriched and lifted to LLM space by a stack of spatially-aware self-attention layers where both queries and keys are rotated by 3D
Rotary Position Embeddings (3D RoPE)~\cite{su2024rope}.
Applying 3D RoPE to both Q and K encodes pairwise relative 3D
distances between voxels directly in the attention scores, giving the projector spatial awareness without positional embedding lookups. The output is $\mathbf{S}\in\mathbb{R}^{N\times D_\text{lm}}$ with LLM hidden dimension $D_\text{lm}$.

\paragraph{Location-Aware Pathway.}
The location-aware pathway produces a single token that encodes the visual content at the queried 3D point. Given the query location
$\mathbf{p} \in \mathbb{R}^{3}$, we identify the closest voxel center (with voxel id \(i\)). The voxel feature $\mathbf{F}_{i}$ is concatenated
with the normalized query location and passed through a four-layer MLP projecting to LLM hidden dimension:
\begin{equation}
    \mathbf{T}_\text{loc}
        = \mathrm{MLP}\!\left([\mathbf{F}_{i}\,;\,
          \mathbf{p}]\right)
        \;\in\mathbb{R}^{D_\text{lm}}.
\end{equation}
The concatenation of raw spatial coordinates with the voxel feature allows the MLP to jointly reason about \emph{what} is at a location and \emph{where} that location is, without a separate conditioning mechanism.
When no query location is provided the pathway is suppressed and no location token is injected. 

\paragraph{Context-Aware Pathway.}
The context-aware pathway produces a compact, prompt-independent scene summary.
A set of $K{=}8$ learnable query tokens (${Q}_\text{learn}$) cross-attends over the sparsified and spatially enriched features $\mathbf{S}$ through two cross-attention layers
\begin{equation}
    \mathbf{T}_\text{ctx}
        = \mathrm{XAttn}(\mathbf{Q}_\text{learn},\,\mathbf{S})
        \;\in\mathbb{R}^{K\times D_\text{lm}}.
\end{equation}
Because the queries are unconditional, they learn to summarize scene-level properties (object layout, room type, spatial extent) useful regardless of the query.  
We observe that this global context allows the task-aware pathway to specialize more sharply: without context tokens, the compressor attends broadly across all voxels; with them, attention concentrates on query-relevant spatial regions.

\paragraph{Task-Aware Pathway.}
The sparsified visual representation is consumed by the task-aware pathway for further sparsification of the tokens with respect to the user task query. 
The language query $L$ is processed by two self-attention layers to $\mathbf{L}_\text{refined}$.
Subsequently, we add a 3D RoPE embedding~\cite{su2024rope} to each voxel token, which then cross-attends to the text tokens.
This asymmetry encodes the relevance of each prompt token for each spatial location in the attention score, allowing selective refinement of spatially relevant voxels.
The update is applied as a residual with zero-initialized output projection for stable training:
\begin{equation}
    \mathbf{S}_\text{task}
        = \mathrm{LayerNorm}\!\left(
            \mathbf{S} + \mathbf{W}_\text{out}
            \cdot \mathrm{XAttn}_\text{3DRoPE}(
                \mathbf{S},\,\mathbf{L}_\text{refined})
          \right).
\end{equation}

The task-refined scene $\mathbf{S}_\text{task}$ is compressed via 128 learnable tokens to $\mathbf{T}_\text{task}$ LLM-ready tokens.

\section{Experiments}

\subsection{Experimental Setup}
\paragraph{Datasets.}
We follow standard training protocols within the scene-centric 3D VLM field \cite{chen2024ll3da,halacheva2025gaussianvlm} and train our models within the ScanNet \cite{dai2017scannet} settings. Notably, we use 3D GS representation \cite{li2025scenesplat} as visual input. We perform joint training on multiple of tasks without task-specific fine-tuning: general question-answering (ScanQA \cite{scanqa_22_cvpr}), situated question-answering (SQA3D \cite{ma2022sqa3d}) and location-grounded captioning (on Nr3D \cite{achlioptas2020referit_3d}, ScanRefer \cite{chen2020scanrefer}).

\paragraph{Evaluation.} Following baseline protocols \cite{halacheva2025gaussianvlm}, we evaluate question-answering on a multitude of metrics covering linguistic similarity (METEOR (M) \cite{banerjee2005meteor}, ROUGE (R) \cite{chin2004rouge}, BLEU-4 (B4) \cite{papineni2002bleu}, CIDEr (C)~\cite{vedantam2015cider}), and exact-match accuracy. For the latter, we adopt the refined EM protocol (EM-R)~\cite{huang2023embodied}. For captioning, we evaluate target object identification via accuracy (Acc) and caption similarity via Sentence-BERT Similarity (Sim) \cite{reimers-gurevych-2019-sentence}, with metrics discussion provided in Supp.  Sec.~\ref{supp_sec:exp_ngram}. Results are reported on the validation sets of ScanRefer, Nr3D, ScanQA and test set of SQA3D.

\paragraph{Training.}
Training is carried out in two stages. In the first stage, only the location-aware pathway is active, with the model being trained on
point-based grounded captioning (Nr3D and ScanRefer), using the single location token as the sole visual input. This
allows the location extractor to learn a stable voxel-to-language mapping before the full scene representation is introduced. 
In the second stage, all three pathways are active and the model is trained jointly on the full task mixture. Both stages are trained under the standard autoregressive language modeling objective (Supp. Sec. \ref{supp_sec:hyperparam}).

\paragraph{Implementation Details.}
P$^3$-VLM uses Qwen3-VL-4B~\cite{bai2025qwen3} as LLM backbone, fine-tuned with LoRA~\cite{hu2022lora} with rank 16 \cite{halacheva2025gaussianvlm,huang2023embodied}.
The 3D encoder is initialized from SceneSplat \cite{li2025scenesplat} and kept unfrozen. The three visual pathway components are trained from scratch. Training takes 10 hours on 2 NVIDIA-H200 GPUs with 5 epochs of pre-training the location pathway, followed by 12 epochs of full model training, using batch size 16 across both stages. We use the AdamW optimizer with a cosine annealing learning rate schedule, learning rate 3e-5 and weight decay of 0.05. We perform evaluation every epoch. For further training and hyperparameter details, please refer to Supp. Sec. \ref{supp_sec:hyperparam}. We train and evaluate two variants of P$^3$-VLM differing solely in the adopted location prompting strategy - bounding box (\textit{P$^3$-VLM (b)} with box format \textit{[center, yaw, extent]}) or point (\textit{P$^3$-VLM (p)}). For the latter we use the centers of the objects, derived from the ground truth segmentation \cite{dai2017scannet}.

To isolate language model performance from architectural contributions, we also retrain one of the baselines, GaussianVLM, using its original training protocol while replacing its language decoder with ours: Qwen3-VL-4B (denoted GaussianVLM$^{\text{Q}}$). This ensures that both P$^3$-VLM and GaussianVLM$^{\text{Q}}$ share identical 3D (SceneSplat) and language backbones.

\subsection{Grounded Object Understanding}
\label{sec:exp_grounded}
We evaluate P$^3$-VLM on the ability to perform grounded visual understanding via location-prompted object captioning on Nr3D and ScanRefer. As other methods do not report class accuracy on the captioned object, we obtain the predictions from all baseline models and extract the predicted object class with an LLM-based evaluator (Gemini 2.5 Flash \cite{comanici2025gemini}). More details on our evaluation protocol are provided in Sec. \ref{supp_sec:evaluator}.

\begin{wraptable}{r}{0.45\textwidth}
    \vspace{-0.5cm}
    \caption{\textbf{Object captioning performance on Nr3D and ScanRefer.} We report classification accuracy (Acc) and caption similarity (Sim). P$^3$-VLM consistently outperforms baselines in Acc, showing superior visual grounding across point (p) and box (b) prompts. GaussianVLM$^{Q}$: \cite{halacheva2025gaussianvlm} with Qwen3-VL-4B LLM, retrained.
    }
    \centering
    \footnotesize
    \begin{tabular}{@{\hskip 0mm}l c c c c }
    \toprule
      & \multicolumn{2}{c}{Nr3D} & \multicolumn{2}{c}{ScanRefer}\\
      \cmidrule{2-3} \cmidrule{4-5}
      Model  & Acc & Sim & Acc & Sim\\
      \midrule
      LL3DA (b) & 47.7 & 48.1 & 48.0 & 55.9\\
      GaussianVLM (b) & 46.2& 48.2 & 50.4 & 59.1 \\
      GaussianVLM$^{Q}$ (b) & 47.1 & 47.8 & 51.7 & 59.3 \\
      \midrule
      P$^3$-VLM (p) & 52.1 & 46.0  & 56.1 & 54.2  \\
      P$^3$-VLM (b) & \textbf{52.4} & 46.3 & \textbf{58.4} & 52.5 \\
       \bottomrule
    \end{tabular}
    \label{tab:nr3dScanref}    
    \vspace{-0.5cm}
\end{wraptable}
\paragraph{Results.} As shown in Table~\ref{tab:nr3dScanref}, both P$^3$-VLM (b) and P$^3$-VLM (p) obtain significantly better class-accuracy on both Nr3D and ScanRefer captioning. 
As expected, P$^3$-VLM (b), our box-based method, slightly outperforms our point-based method on in-domain grounded tasks, courtesy of having access to the object's geometric properties. 
The stable sentence similarity results (Sim) demonstrate that P$^3$-VLM not only identifies the correct object but also maintains high linguistic similarity to the entire caption.

\paragraph{Priors analysis in points and boxes.}
Even point prompts are not entirely free of priors: absolute coordinates can still correlate with object category through scene-layout regularities. We examine and quantify these residual priors with a geometry-only kNN classifier, trained on the Nr3D train split and evaluated on the validation split (38 classes with at least 8 instances). Using the object center as the point prompt, the horizontal position ($xy$) alone yields $8.1\%$ accuracy, the height ($z$) alone $17.8\%$, and the full 3D center ($xyz$) $16.5\%$, compared to $33.0\%$ for the full bounding box. Height thus accounts for nearly all of the residual signal, while absolute room position is a weak cue. Per class, the center-based kNN reaches $\geq30\%$ recall on only 3/38 classes (chair, picture, monitor), compared to 15/38 for the box-based one: the residual prior is confined to objects with a consistent mounting or placement height, whereas shape-distinctive objects (e.g., bed, keyboard, door) leak almost entirely through box geometry.

\begin{table*}[t]
    \centering
    \begin{minipage}[t]{0.45\textwidth}
        \centering
        
        \caption{\textbf{Robustness of 3D VLMs using box prompting to geometric shortcut disruption.} Class accuracy (\%) on Nr3D under two failure regimes: non-tight bounding boxes, and noisy reconstructions, provided together with
        In-distribution performance for reference. Models relying on
        bounding box semantic leakage degrade sharply across both conditions.}
    
        \centering
            \footnotesize
            \begin{tabular}{@{\hskip 0mm}l | c | c c}
             \toprule
                & In-Dist. & \makecell{Non-Tight\\Boxes} & \makecell{Noisy\\Recon.} \\
            \midrule
                LL3DA                & 47.7 & 28.6 
                & 33.4 \\
                GaussianVLM          & 46.2 & 25.3 
                & 32.2 \\
            \midrule
               \textbf{P$^3$-VLM (b)}          
               & \textbf{52.4} 
               & \textbf{36.4} 
               & \textbf{39.9} \\
            \bottomrule
            \end{tabular}
            \label{tab:ood_grounding}
    \end{minipage}
    \hfill
    \begin{minipage}[t]{0.53\textwidth}
        \centering
        \caption{\textbf{P$^3$-VLM (p) robustness to query point perturbation.} We train on object centers (query offset $x{=}0$) and evaluate with query points drawn uniformly at random from a ball of radius $x \times$ the box extent around the center, or uniformly from the object surface. We observe graceful degradation even at extreme offsets.}
        \centering
        \footnotesize
        \begin{tabular}{@{\hskip 0mm}l@{\hskip 0.7mm} | c c c c }
        \toprule
          Query& \multicolumn{2}{c}{Nr3D} & \multicolumn{2}{c}{ScanRefer}\\
          \cmidrule{2-3} \cmidrule{4-5}
           Offset  & Acc & Sim & Acc & Sim\\
        \midrule
          $0$ & 52.1 & 46.0  & 56.1& 54.2  \\
          \midrule
          $0.1$ 
            & 50.6 {\textcolor{red!70!black}{(-1.5)}} 
            & 45.6 {\textcolor{red!70!black}{(-0.4)}} 
            & 56.5 {\textcolor{red!70!black}{(+0.4)}} 
            & 54.0 {\textcolor{red!70!black}{(-0.2)}}\\
          $0.3$ 
            & 49.9 {\textcolor{red!70!black}{(-2.2)}} 
            & 45.7 {\textcolor{red!70!black}{(-0.3)}} 
            & 56.1 {\textcolor{red!70!black}{(0.0)}} 
            & 54.0 {\textcolor{red!70!black}{(-0.2)}}\\
          $0.6$ 
            & 47.4 {\textcolor{red!70!black}{(-4.7)}} 
            & 45.5 {\textcolor{red!70!black}{(-0.5)}} 
            & 54.1 {\textcolor{red!70!black}{(-2.0)}} 
            & 53.6 {\textcolor{red!70!black}{(-0.6)}} \\
          $0.9$ 
            & 46.6 {\textcolor{red!70!black}{(-5.5)}} 
            & 45.1 {\textcolor{red!70!black}{(-0.9)}} 
            & 52.8 {\textcolor{red!70!black}{(-3.3)}} 
            & 53.1 {\textcolor{red!70!black}{(-1.1)}}\\
        \midrule
          Surf.
            & 50.7 {\textcolor{red!70!black}{(-1.4)}}
            & 45.4 {\textcolor{red!70!black}{(-0.6)}}
            & 56.9 {\textcolor{red!70!black}{(+0.8)}}
            & 54.1 {\textcolor{red!70!black}{(-0.1)}}\\
           \bottomrule        
           \end{tabular}
        \label{tab:ood_grounding_points}
    \end{minipage}
\end{table*}

\label{sec:exp_ood}
\subsection{Out-of-Distribution Generalization}
\paragraph{Robustness to Bounding Box Perturbation.}
The bounding boxes provide a categorical prior that often does not hold in noisy or out-of-distribution data, e.g., spaces with larger cabinets or larger windows. 
We evaluate two failure regimes. (1) Non-tight bounding boxes, produced by imperfect detectors or coarse annotations, dilute the geometric signal that the model has come to depend on. (2) Noisy or partial reconstructions introduce geometric artifacts that corrupt box estimates further. 

To simulate regime (1), we shrink each GT box by a per-class scale factor $s_c = \bar{\iota}_c^{1/3}$, where $\bar{\iota}_c$ is the mean IoU of Vote2Cap-DETR \cite{chen2023end} detections on class $c$, so that the resulting box volume matches the average detection quality of a state-of-the-art detector. Vote2Cap-DETR is chosen as it is adopted within the LL3DA baseline, keeping the simulation as faithful as possible to the conditions these models were designed for. 
Regime (2) is simulated symmetrically: boxes are expanded by the same per-class factor $1/s_c$, reflecting the tendency of detectors to produce loose predictions under reconstruction noise. Full per-class IoU values and implementation details are provided in Supp. Sec. \ref{supp_sec:ood_grounding}. As shown in Tab. \ref{tab:ood_grounding}, each of these conditions induces a severe performance drop -- not because the visual scene has changed dramatically, but because the leakage channel the model relied on has been disrupted. 
Notably, P$^3$-VLM (b) remains the most stable.

\begin{wraptable}{r}{0.45\textwidth}
\vspace{-0.5cm}
    \caption{ScanQA Results. GaussianVLM$^{Q}$: \cite{halacheva2025gaussianvlm} with Qwen3-VL-4B LLM, retrained.}
    \centering
    \footnotesize
    \begin{tabular}{l c c c}
    \toprule
      \multicolumn{4}{c}{ScanQA} \\
      \midrule
      Model & EM-R &  M & R\\
      \midrule
      
      ScanQA \cite{scanqa_22_cvpr}& / &13.14&  33.33 \\
      3D-VLP\cite{3dvlp} & / &13.53 &34.51 \\
      VoteNet+MCAN \cite{yu2019deep}& / & 11.4 &29.8 \\
      LSceneLLM \cite{zhi2024lscenellm}& / & 17.95 & /\\
      LL3DA \cite{chen2024ll3da}& 30.3 &22.8 &34.7\\
      GaussianVLM \cite{halacheva2025gaussianvlm}& 30.7&  \textbf{22.9} & 34.8\\
      GaussianVLM$^Q$ & 30.9 & 21.1 & \textbf{35.3}\\
      \midrule
      \textbf{P$^3$-VLM (p)} &  \textbf{31.9}  & 22.6 & \textbf{35.3} \\
      \textbf{P$^3$-VLM (b)} & 30.5 & 21.2 & 33.9 \\
       \bottomrule
    \end{tabular}
    \label{tab:scanqa}
\end{wraptable}
\paragraph{Robustness to Query Point Perturbation.}
To test the spatial stability of P$^3$-VLM (p), trained on exact center coordinates, we evaluate it on random query points: for offset ratio $x \in \{0.1, 0.3, 0.6, 0.9\}$, the query is drawn uniformly at random from a ball of radius $x$ times the box extent, centered at the object center. At $x{=}0.9$ the sampling region covers nearly the full object volume. As shown in Tab.~\ref{tab:ood_grounding_points}, performance degrades gracefully with perturbation distance, dropping by only 5.5 accuracy points on Nr3D and 3.3 on ScanRefer at the most extreme setting ($0.9\times$ extent). We further query points sampled uniformly from the object \emph{surface}, which changes performance only modestly, confirming that P$^3$-VLM (p) does not depend on the privileged center point.

\begin{figure}
    \centering
    \includegraphics[width=\linewidth]{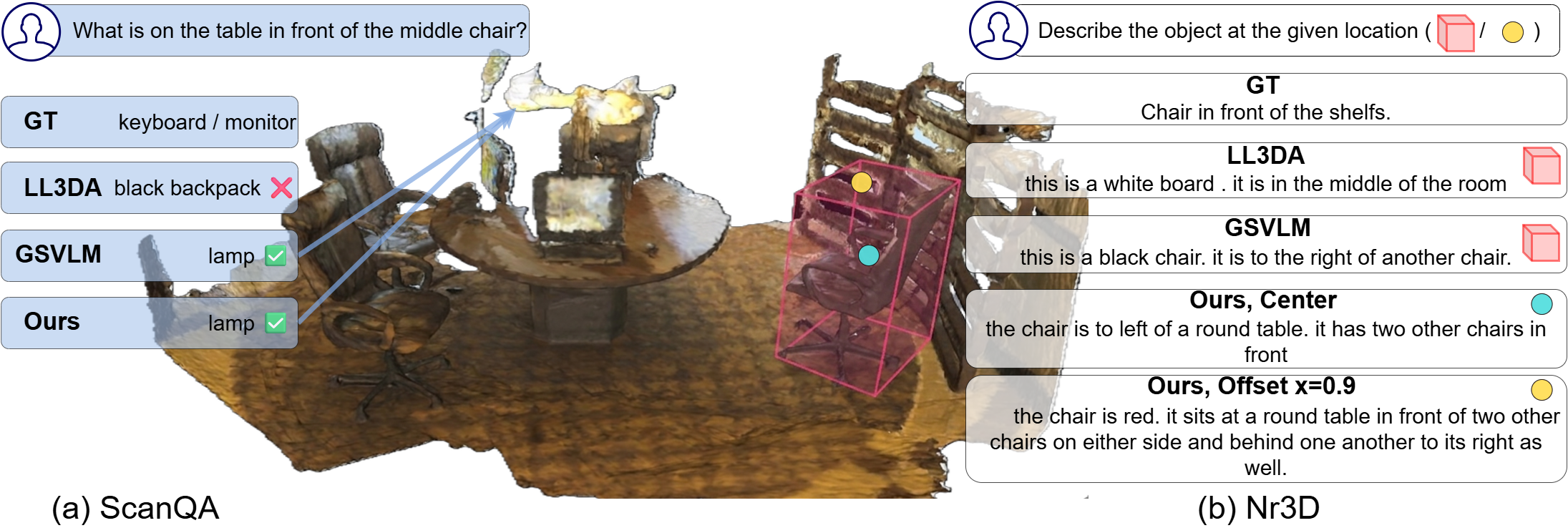}
    \vspace{-15pt}
    \caption{\textbf{Qualitative results on ScanQA and Nr3D.} (a) P$^3$-VLM can correctly answer scan questions, sometimes even with answers not captured by the benchmark ground truth.  (b) Through the combination of task, context, and location pathway, on Nr3D, P$^3$-VLM identifies the correct object class, surrounding context, and spatial relationships. The performance remains robust even under perturbations (offset factor $x=0.9$).}
    \label{fig:quali1}
\end{figure}

\subsection{Ungrounded Scene Understanding}
We further examine the generalist capabilities of P$^3$-VLM on three ungrounded scene understanding benchmarks: ScanQA, SQA3D, and GaussianVLM out-of-distribution benchmark (GaussianVLM-OOD) \cite{halacheva2025gaussianvlm} (Tables~\ref{tab:scanqa},~\ref{tab:sqa3d},~\ref{tab:ood}). The latter probes VLM generalization on an object counting task on out-of-distribution ScanNet++ scenes \cite{yeshwanth2023scannet++}, measured via exact match accuracy (EM).
P$^3$-VLM (p) achieves state-of-the-art performance across all three benchmarks, surpassing the state-of-the-art GaussianVLM on most metrics despite being trained with the strictly harder point-based prompting protocol.
On the ungrounded scene-level OOD benchmark, P$^3$-VLM (p) reaches 17.31 EM versus 9.3 for GaussianVLM, with P$^3$-VLM (b) scoring 15.42. The +1.89 EM advantage of (p) over (b) on this benchmark indicates that removing bounding box supervision improves out-of-distribution robustness, consistent with our hypothesis that point-based training discourages reliance on box-derived shortcuts and encourages stronger visual representations.

\begin{table}[t]
\centering
\begin{minipage}[t]{0.62\linewidth}
    \caption{SQA3D Results. \\GaussianVLM$^{Q}$: \cite{halacheva2025gaussianvlm} with Qwen3-VL-4B LLM, retrained.}
    \centering
    \small
    \vspace{0pt}
    \begin{tabular}{l c c c c c}
    \toprule
      \multicolumn{6}{c}{SQA3D}\\
      \midrule
      Model & EM-R & B4 & C & M & R\\
      \midrule
      LL3DA \cite{chen2024ll3da}& 43.3 & 1.0 & 107.2 & 21.4 & 42.14 \\
      LEO \cite{huang2023embodied} & 50.9 & 9.4 & 124.7 & 25.5 & 48.4\\
      GaussianVLM \cite{halacheva2025gaussianvlm}& 51.4 & 17.1 & 129.6 & {26.4} & 50.2 \\
      GaussianVLM$^{Q}$ & \textbf{51.9} & 18.9 & 125.4 & \textbf{26.7} & 50.0 \\
      \midrule
      \textbf{P$^3$-VLM (p)} & \underline{51.8} & \textbf{30.6} & \textbf{130.4} & \underline{26.6} & \underline{50.3}\\
      \textbf{P$^3$-VLM (b)} & \textbf{51.9} & \underline{19.5} & \underline{130.3} & {26.4} & \textbf{50.5}\\
      \bottomrule
    \end{tabular}
    \label{tab:sqa3d}
\end{minipage}
\hfill
\begin{minipage}[t]{0.35\linewidth}
    \caption{Out-of-distribution results. Zero-shot evaluation on scenes from unseen ScanNet++ \cite{yeshwanth2023scannet++}.}
    \centering
    \small
    \vspace{0pt}
    \begin{tabular}{l c}
    \toprule
      \multicolumn{2}{c}{GaussianVLM-OOD}\\
      \midrule
      Model & EM\\
      \midrule
      LL3DA \cite{chen2024ll3da}& 1.5\\
      LEO \cite{huang2023embodied}& 7.9 \\
      GaussianVLM \cite{halacheva2025gaussianvlm}& 9.3 \\
      \midrule
      \textbf{P$^3$-VLM (p)} & \textbf{17.31} \\
     \textbf{ P$^3$-VLM (b)} & 15.42\\
      \bottomrule
    \end{tabular}
    \label{tab:ood}
\end{minipage}
\end{table}
\vspace{-0.1cm}
\subsection{Ablation Studies} 
\label{sec:ablations}
We evaluate the contribution of different components within P$^3$-VLM through a series of ablation studies considering our tri-pathway design (Tab. \ref{tab:ablations}) with ablations on the design of separate components provided in Supp. Sec. \ref{supp_sec:ablations2}. Our ablations cover location-grounded and ungrounded tasks with both object-level and scene-level understanding focus.

We show the necessity of understanding 3D scenes at three levels of granularity by ablating the three pathways of P$^3$-VLM. Removing any pathway causes consistent drops in task performance. Notably, the location pathway proves an important component enabling scene-level tasks (SQA3D), although it is used only with location-based queries. As we observe, it allows the other two pathways to focus on scene-level understanding rather than generalist capabilities.

\paragraph{Matched baseline.}
To fully isolate the contribution of our visual processing, we construct a \emph{matched baseline} that shares the 3D encoder and LLM with P$^3$-VLM, but uses more simplistic visual features: the query point is represented by the raw feature of its nearest voxel, and scene information is provided only via the context tokens, i.e., without our location MLP and without the task pathway. The matched baseline falls far behind P$^3$-VLM on point-based grounding (Nr3D: $39.8\%$ vs.\ $52.1\%$, ScanRefer: $44.2\%$ vs.\ $56.1\%$ Acc), showing that the pathway choices and their design are well-motivated.

\begin{table}[t]
    \caption{\textbf{Design ablation studies on P$^3$-VLM (p).} Top: Each of the pathways is removed independently. Bottom: A matched baseline that isolates visual processing innovations by employing a multi-pathway variant with minimal structural complexity -- utilizing only nearest-voxel features at the query point and global context tokens.
    }
    \centering
    \small
    \begin{tabular}{@{\hskip 0mm}l | c c | c c c | c c c c c }
    \toprule
        & \multicolumn{2}{c|}{Nr3D}
        & \multicolumn{3}{c|}{ScanQA}
        & \multicolumn{5}{c}{SQA3D} \\
        \cmidrule{2-3} \cmidrule{4-6} \cmidrule{7-11}
        & Acc & Sim & EM-R & M & R & EM-R & B4 & C & M& R \\
    
    \midrule
     w/o location pathway  & 49.3& 43.9& 30.4 & 24.2 & 34.1 & 45.6 & 0 & 114.6 & 23.4 & 44.3 \\
     w/o task pathway      & 50.2 & 
     46.0& 30.1& 24.2 & 19.2 & 47.9 & 22.5 & 119.0 & 24.1 & 46.2 \\
     w/o context pathway   & 50.2 & 45.2 &  22.7 &14.7 & 23.3 & 49.4 & 34.8 & 123.2 & 25.2 & 47.8 \\ 
    \midrule
     Matched baseline      & 39.8 & 41.9 & 31.3 & 21.7 & 33.9 & 49.6 & 20.2 & 124.2 & 25.3 & 48.1 \\
    \midrule
    \textbf{Full model}   & 52.1 &46.0 & 31.9 & 22.9& 34.8 & 51.8 &30.58 &130.4 &26.59 &50.32\\
    \bottomrule
    \end{tabular}
    \label{tab:ablations}
    \vspace{-10pt}
\end{table}

\section{Conclusions}
In this work, we highlight the problem of semantic leakage in current 3D VLMs, causing these models to neglect visual information.
We propose a new point-based grounding prompting protocol and show that other methods struggle under this new setting. 
To address this, we present P$^3$-VLM, a novel 3D VLM based on a tri-pathway design.
We demonstrate the proposed point-based grounding approach with our architecture leads to improved visual understanding, demonstrated through improved performance, generalization, and robustness on multiple 3D tasks and benchmarks.

\paragraph{Limitations}
First, P$^3$-VLM requires 3D GS representations as input, introducing an extra reconstruction step that adds complexity; however, this step has been performed at
scale in prior works~\cite{li2025scenesplat,ma2025scenesplat++} and industrial applications \cite{pix4dcatch_2025,aws_guidance_2024}, and is shown to yield richer visual features \cite{halacheva2025gaussianvlm}.
Second, our evaluation is primarily focused on perception-oriented benchmarks, with limited coverage of language-intensive or dialogue-driven tasks. Extending P$^3$-VLM to more complex multimodal reasoning settings remains an important direction for future work.
Third, our approach relies on voxelized visual features, currently obtained via the SceneSplat encoder. Evaluating how well the method generalizes to alternative 3D visual encoders is an interesting avenue for further investigation. 

\paragraph{Impact}
We envision this work as a step toward enabling real-world applications such as high-level planning for navigation and visual assistance systems. Additionally, we hope it draws attention to the issue of information leakage in 3D grounding, encouraging the development of more robust and general-purpose 3D vision-language models.

\clearpage

\section*{Acknowledgments}
This research was partially funded by the Ministry of Education and Science of Bulgaria (support for INSAIT, part of the Bulgarian National Roadmap for Research Infrastructure).

{\small
\bibliographystyle{plain}
\bibliography{bibliography}
}

\appendix
\clearpage

\section{Experimental Details}
\label{supp_sec:exp}

\subsection{Experimental Details: Point-Based Grounding Adaptation}
\label{supp_sec:exp_tunePoint}

This section provides full experimental details for the results 
reported in Sec.~\ref{sec:pointVsBox}.

\paragraph{Blind LLM Experiments.}
We fine-tune  Qwen3-VL-4B-Instruct \cite{bai2025qwen3} on object classification within the Nr3D validation set under two prompt conditions: bounding box and point location. In both cases, the location query is provided as a text string with coordinates rounded to two decimal places. Bounding boxes are encoded via center coordinates, yaw angle and scale, as $\langle c_x, c_y, c_z, yaw, w, h, l \rangle$ without explicitly naming the format in the prompt. Point prompts contain only the 3D coordinate $\langle x, y, z \rangle$.

Fine-tuning is performed with LoRA using rank $r{=}16$, scaling factor $\alpha{=}32$, dropout 0.05, and a learning rate of $3 \times 10^{-5}$. The LoRA adapters are applied to the query, key, value, and output projection layers (\texttt{q\_proj}, \texttt{k\_proj}, \texttt{v\_proj}, \texttt{o\_proj}). As shown in Fig.~\ref{fig:blind_llm_curve}, 
validation accuracy saturates between epochs 6--8 and overfitting emerges thereafter.

\paragraph{Scene-Centric VLMs.}
Based on the convergence behavior observed for the blind LLM, we fine-tune all scene-centric VLMs for 6 epochs. To preserve learned visual representations, only the LLM component of each model is fine-tuned, using the same LoRA configuration described above. Fine-tuning is performed exclusively on Nr3D. We evaluate after every epoch and report results from the checkpoint with the highest validation accuracy. Point prompt format is consistent with the blind LLM  and P$^3$-VLM (p) experiments ($\langle x, y, z \rangle$). For bounding box prompting evaluation, we adhere to the bounding box formats used by the specific baselines.

\begin{figure}[b]
    \centering
    \includegraphics[width=0.8\linewidth]{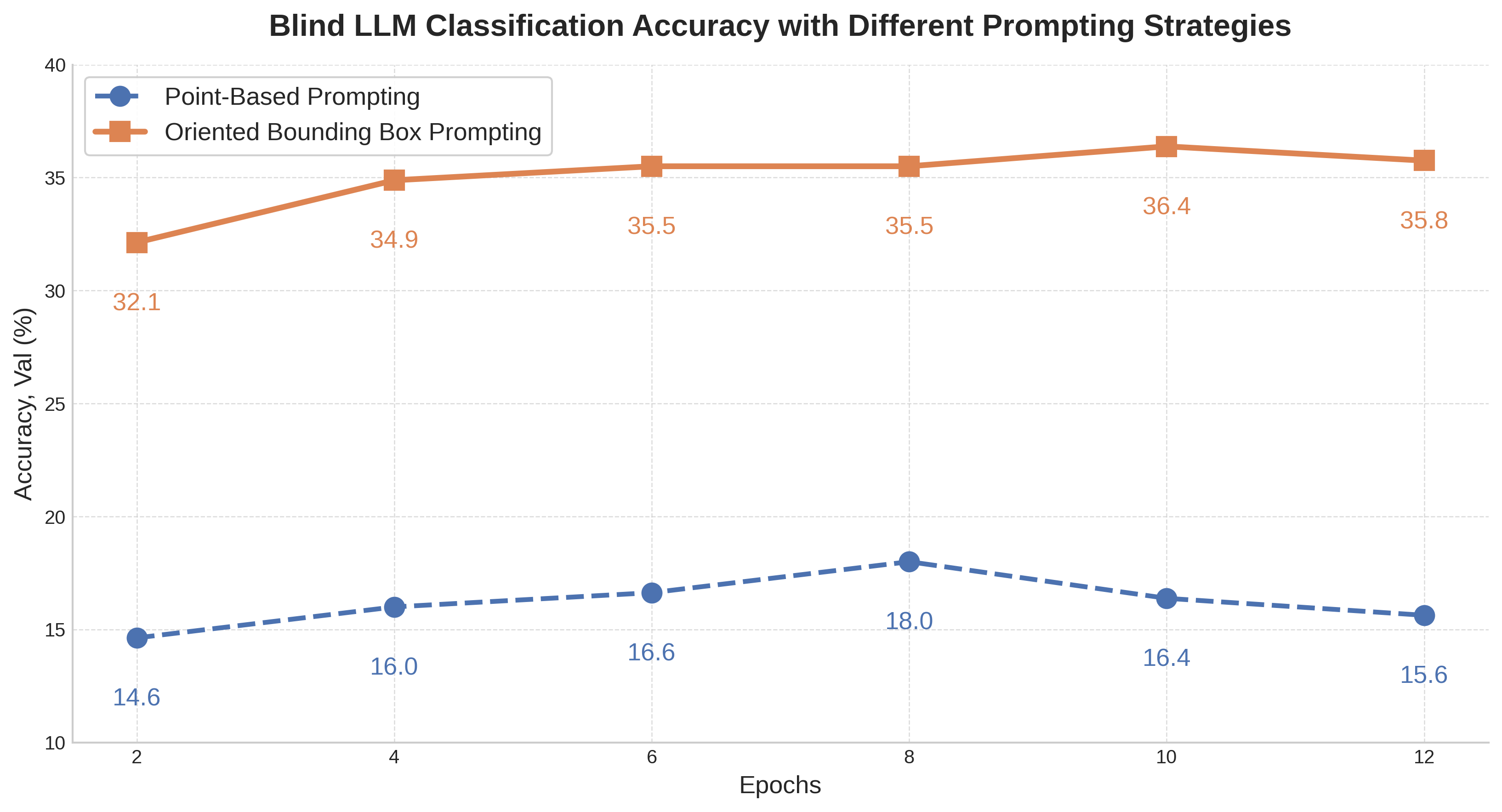}
    \caption{\textbf{Validation accuracy of the blind LLM during fine-tuning under bounding-box and point-based prompting.} 
    Performance saturates between epochs 6--8, with overfitting emerging afterwards. This curve motivates the 6-epoch fine-tuning budget used for all VLM experiments.}
    \label{fig:blind_llm_curve}
\end{figure}

\subsection{N-gram Metrics Conceal Failures in Spatial Understanding}
\label{supp_sec:exp_ngram}

Standard 3D scene understanding benchmarks such as ScanRefer and Nr3D evaluate dense captioning models using n-gram overlap metrics -- CIDEr, METEOR, ROUGE -- gated by an IoU threshold of 0.25 or 0.50. While these metrics are practical for large-scale evaluation, they measure linguistic surface similarity rather than semantic correctness, and in particular are insensitive to whether the model has correctly identified the class of the referred object. A response that fluently describes the wrong object category can achieve a high n-gram score if it shares common words with the reference caption \cite{halacheva2025gaussianvlm}.

To quantify this effect, we compare the correlation of standard n-gram evaluation with class accuracy: the fraction of generated captions from which the correct object category can be extracted. This metric directly measures whether the model has understood what it is looking at, independently of linguistic fluency. As shown in Tab. \ref{tab:ngram_vs_acc}, models that achieve competitive n-gram scores exhibit strikingly low class accuracy -- in several cases falling below a blind LLM that receives no visual input whatsoever. These results indicate that n-gram metrics systematically overestimate the spatial understanding of current 3D VLMs, and motivate the use of class accuracy as a more diagnostic evaluation criterion.

\begin{table}[]
    \centering
    \caption{Different evaluation metrics on captioning on the Nr3D dataset. It is evident that high n-gram based metrics do not signal correct object identification.}
    \begin{tabular}{l c c c c }
    \toprule
       Model  & Sim &  M & R & Class Acc \\
       \midrule
       Blind LLM& 34.8  & 17.9 & 21.3 & 36.4 \\
       LL3DA &48.1 & 5.8 &9.9 & 47.7\\
       GaussianVLM & 48.2 &	20.8  & 48.2 & 46.2\\
       P$^3$-VLM & 46.0 &18.6 & 16.4  & 52.1\\
       \bottomrule
    \end{tabular}
    
    \label{tab:ngram_vs_acc}
\end{table}

\subsection{LLM-Based Class Accuracy Evaluator}
\label{supp_sec:evaluator}
Class accuracy is computed by extracting the primary object class from each generated caption with Gemini 2.5 Flash~\cite{comanici2025gemini} and comparing it to the ground-truth class. Captions are processed in numbered batches with the following prompt:
\begin{quote}
\small\itshape
You are an object classification assistant. You will receive a numbered list of short scene descriptions. For each description, identify the single primary object being described. Reply ONLY with a JSON array of strings -- one object class per entry, in the same order. Use simple, lowercase class names (e.g. "chair", "trash can", "table", "sofa"). If the caption is empty or doesn't describe a recognizable object, reply with "none" for that entry. No explanations, no extra keys, no markdown fences, just the raw JSON array.
\end{quote}
Tab.~\ref{tab:evaluator_examples} shows representative extractions. We further carried out a human quality control on 5\% of the validation predictions and found no extraction errors, owing to the simple nature of the task: the generated captions name the described object explicitly.

\begin{table}[h]
    \centering
    \caption{\textbf{Examples of LLM-based class extraction} on ScanRefer and Nr3D captions generated by LL3DA 3D VLM.}
    \small
    \begin{tabular}{@{\hskip 0mm}l p{0.52\linewidth} c c}
    \toprule
     & Generated caption & Extracted class & GT class \\
    \midrule
    ScanRefer & sos this is a white kitchen cabinet . it is to the left of the oven . eos & kitchen cabinet & kitchen cabinet \\
    ScanRefer & sos there is a rectangular soap dispenser . it is on the wall above the sink . eos & soap dispenser & soap dispenser \\
    Nr3D & sos the stool is in the corner of the room . it is to the right of the whiteboard . eos & stool & stool \\
    Nr3D & sos the plant is on the floor . it is to the right of the door . eos & plant & plant \\
    \bottomrule
    \end{tabular}
    \label{tab:evaluator_examples}
\end{table}

\subsection{Out-of-Distribution Generalization}
\label{supp_sec:ood_grounding}
\paragraph{Robustness to Bounding Box Perturbation.} To obtain our bounding box perturbation scales, we firstly obtain bounding boxes for every ScanRefer and Nr3D instance via the Vote2Cap-DETR object detector. Here we utilize only the validation split of the datasets as the object detector itself has been trained on the ScanNet dataset train split. The scales are derived based on mean IoU between the ground truth boxes and the predicted ones following our approach introduced in Sec. \ref{sec:exp_ood}. This results to the following scales given in Fig. \ref{fig:iou}.
\noindent

\begin{figure}
    \centering
    \includegraphics[width=\linewidth]{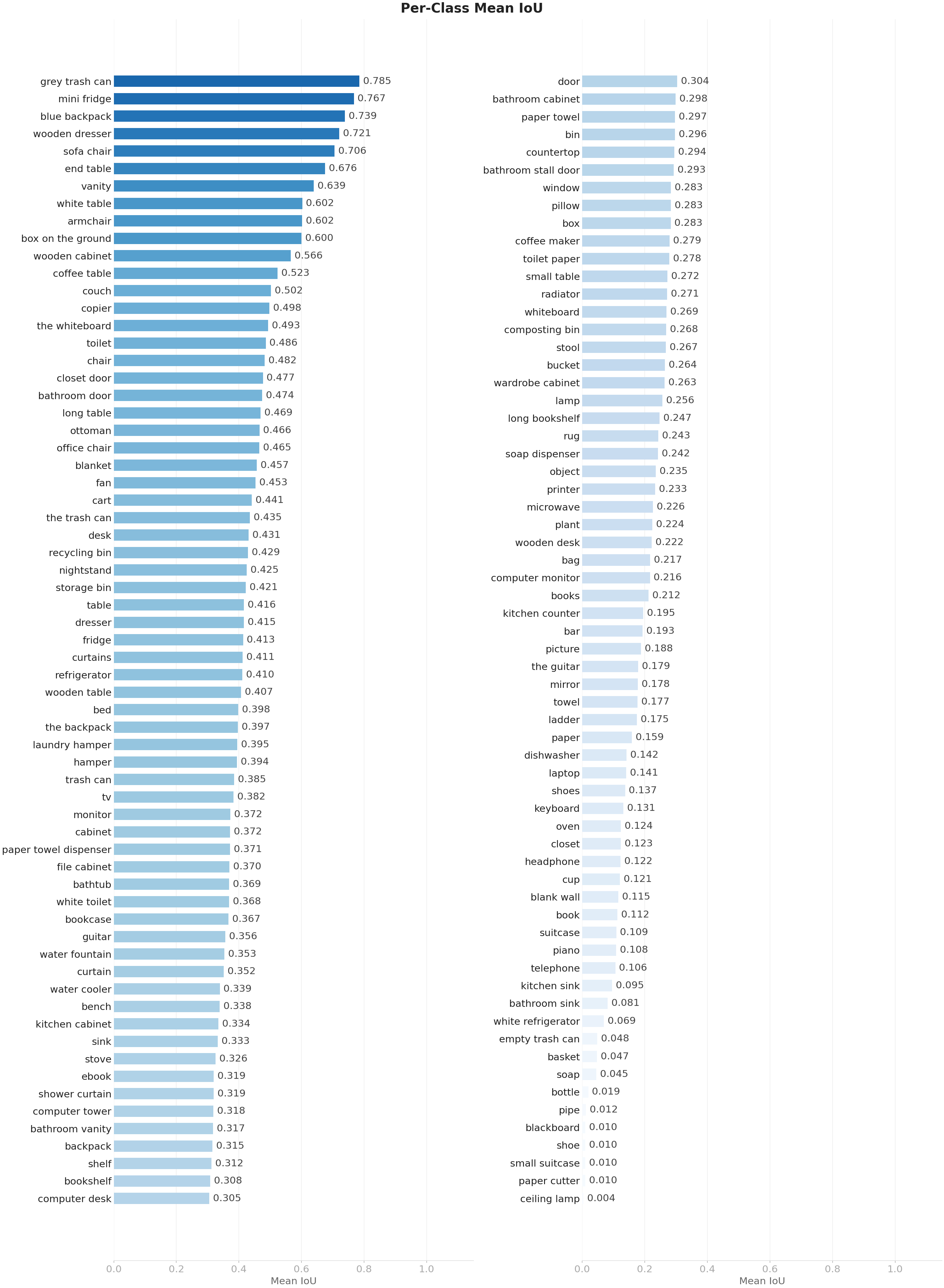}
    \caption{Mean IoU per object class from ScanRefer and Nr3D annotations, validation split.}
    \label{fig:iou}
\end{figure}


\subsection{Implementation Details}
\label{supp_sec:hyperparam}
We train P$^3$-VLM using AdamW with a learning rate of $3\times10^{-5}$,  $\beta = (0.9, 0.999)$, and weight decay $0.05$, with a linear warmup 200 steps followed by cosine decay. Training uses a batch size of 16 with 4 gradient accumulation steps (effective batch size 64) and gradient clipping at norm 5.0. The language backbone is Qwen3-VL-4B-Instruct, fine-tuned via LoRA (rank 16, $\alpha=32$, dropout 0.05) applied to all attention projection 
matrices. The maximum input and output context lengths are set to 2048 and 1024 tokens, respectively. Hyperparameter choices are detailed in Tab. \ref{tab:hyperparams_supp} and are selected in close adherence to the GaussianVLM baseline \cite{halacheva2025gaussianvlm}.
The training objective is the standard autoregressive objective:
\begin{equation}
    \mathcal{L} = -\sum_{t} \log p_{\theta}\left(y_t \mid y_{<t}, T_{loc}, T_{task}, T_{ctx}, L\right)
\end{equation}
, where $y_t$ denotes the $t$-th token of the target response $y = (y_1, \ldots, y_T)$,~$y_{<t}$ the tokens generated earlier, and $\theta$ the set of trainable parameters. 
The loss minimizes the negative log-likelihood of the target sequence conditioned on the language prompt $L$ and all three visual pathway outputs: the location token $\mathbf{T}_\text{loc}$, the compressed task tokens $\mathbf{T}_\text{task}$, and the context summary tokens $\mathbf{T}_\text{ctx}$.
\begin{wrapfigure}{r}{0.4\linewidth}
    \centering
    \includegraphics[width=\linewidth]{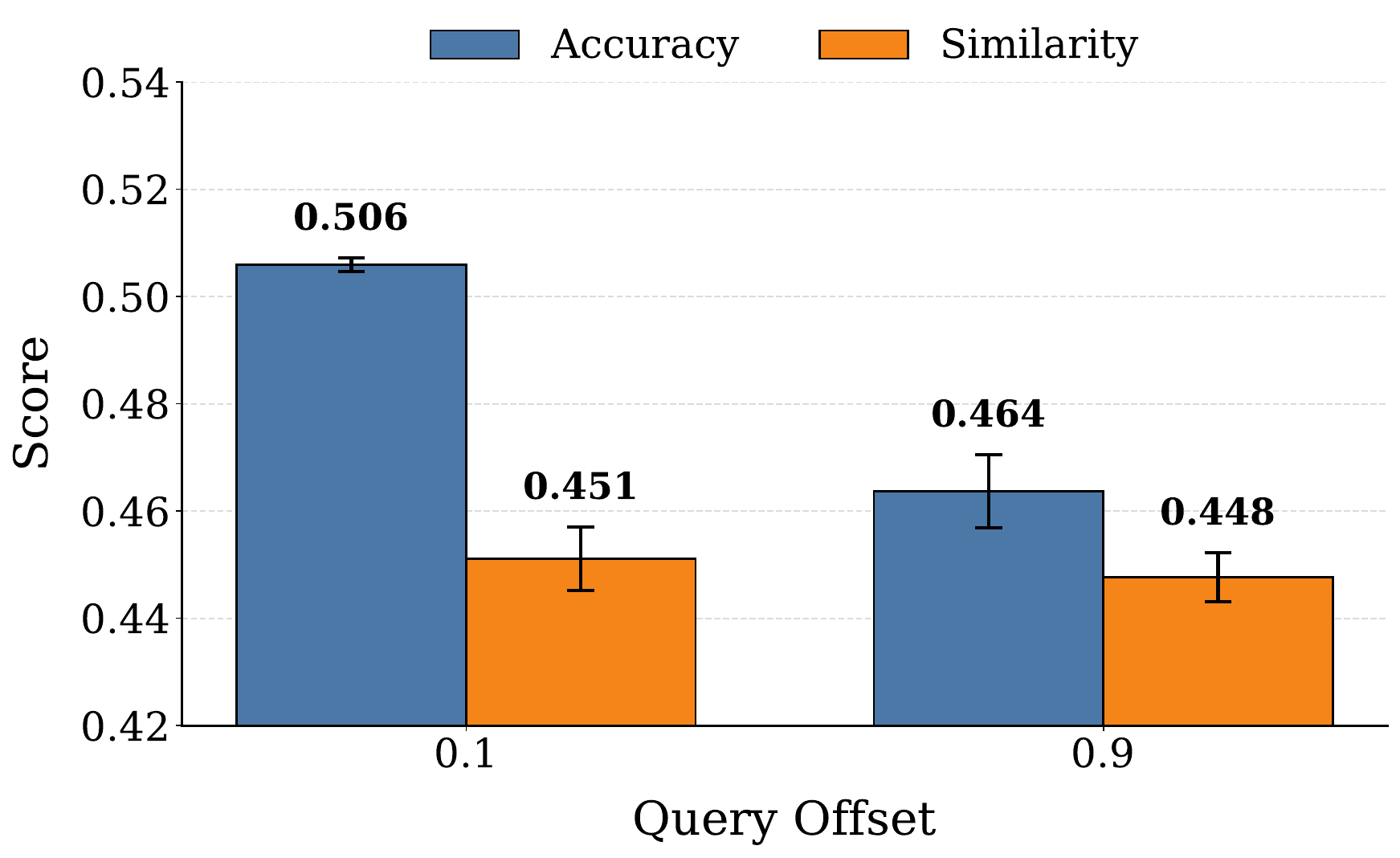}
    \caption{P$^3$-VLM on Nr3D benchmark}
    \label{fig:error_plot}
\vspace{-1.5cm}
\end{wrapfigure}
\subsection{Query point robustness}
We carry out robustness experiments of P$^3$-VLM to the query point offset from the center of the object. 
We notice that P$^3$-VLM is very stable to even the extreme condition where the query-point lies near the edge of the object. We plot the mean accuracy across the different samples with error bar indicating $1\sigma$ deviation, shown in Fig.~\ref{fig:error_plot}. 

\subsection{Training with Surface-Point Grounding}
\label{supp_sec:surface_training}
We acknowledge the center point being a canonical point: its height is a stable, category-specific statistic. To escape this prior, we also analyze prompting with \emph{random surface points} instead. A point sampled uniformly on the object surface may lie anywhere within the object's vertical extent, so its height no longer encodes a consistent per-category value. We retrain P$^3$-VLM with random surface points as the grounding signal during training. As shown in Tab.~\ref{tab:surface_training}, this yields a further, substantial improvement in grounding accuracy (Nr3D: $58.8\%$ vs.\ $52.1\%$ for center training; ScanRefer: $64.3\%$ vs.\ $56.1\%$), while the caption similarity remains comparable. A surface-trained model evaluated on center points performs worse ($50.1\%$ on Nr3D), as the center lies outside of its training distribution.

\begin{table}[h]
    \centering
    \caption{\textbf{Center- vs.\ surface-point grounding} on Nr3D and ScanRefer. c: object center, sp: random surface point. Training on surface points substantially improves class accuracy.}
    \small
    \begin{tabular}{l l l c c c c}
    \toprule
      & & & \multicolumn{2}{c}{Nr3D} & \multicolumn{2}{c}{ScanRefer}\\
      \cmidrule{4-5} \cmidrule{6-7}
      Model & Train & Eval & Acc & Sim & Acc & Sim\\
      \midrule
      LL3DA & box & box & 47.7 & 48.1 & 48.0 & 55.9\\
      GaussianVLM & box & box & 46.2 & 48.2 & 50.4 & 59.1 \\
      P$^3$-VLM (b) & box & box & 52.4 & 46.3 & 58.4 & 52.5 \\
      \midrule
      P$^3$-VLM (p) & c & c & 52.1 & 46.0 & 56.1 & 54.2 \\
      P$^3$-VLM (p) & c & sp & 50.7 & 45.4 & 56.9 & 54.1 \\
      P$^3$-VLM (p) & sp & c & 50.1 & 42.7 & 50.4 & 50.2 \\
      P$^3$-VLM (p) & sp & sp & \textbf{58.8} & 47.2 & \textbf{64.3} & 54.7 \\
      \bottomrule
    \end{tabular}
    \label{tab:surface_training}
\end{table}

\paragraph{Per-class analysis.}
To understand where the gain originates, we relate the per-class change in accuracy (center- to surface-point training) to the physical object size, computed from the ground-truth 3D bounding boxes. Medium-to-large furniture (tables, desks, cabinets, shelves, bookshelves, seating) improves consistently, with an average gain of 21.3 accuracy points. Small or thin objects (pillows, boxes, monitors, keyboards, cups) improve much less, by 6.2 points on average; this difference is statistically significant (Mann-Whitney U test, $p = 0.014$).
We attribute this behavior to the feature pooling at the query point. Prompting a large object by its center forces the visual features to be pooled from deep inside the object's interior, or even from empty space, e.g., below a table top or inside a cabinet, yielding uninformative features. Surface points, by contrast, anchor the feature pooling directly to regions rich in visible geometry and texture, which substantially improves visual grounding for larger structures.

\section{Qualitative Examples}
We provide qualitative results showcasing the capabilities of P$^{3}$-VLM in Fig. \ref{fig:quali1}, \ref{fig:quali2} across the tasks of 3D grounded captioning (Nr3D benchmark, validation split), general question-anwering (ScanQA benchmark, validation split), situated question answering (SQA3D benchmark, test split) and out-of-distribution generalization (GaussianVLM-OOD benchmark, test split). We observe visual understanding on in-domain settings within ScanNet input scenes, and out-of-domain on ScanNet++.

\begin{figure}
    \centering
    \includegraphics[width=\linewidth]{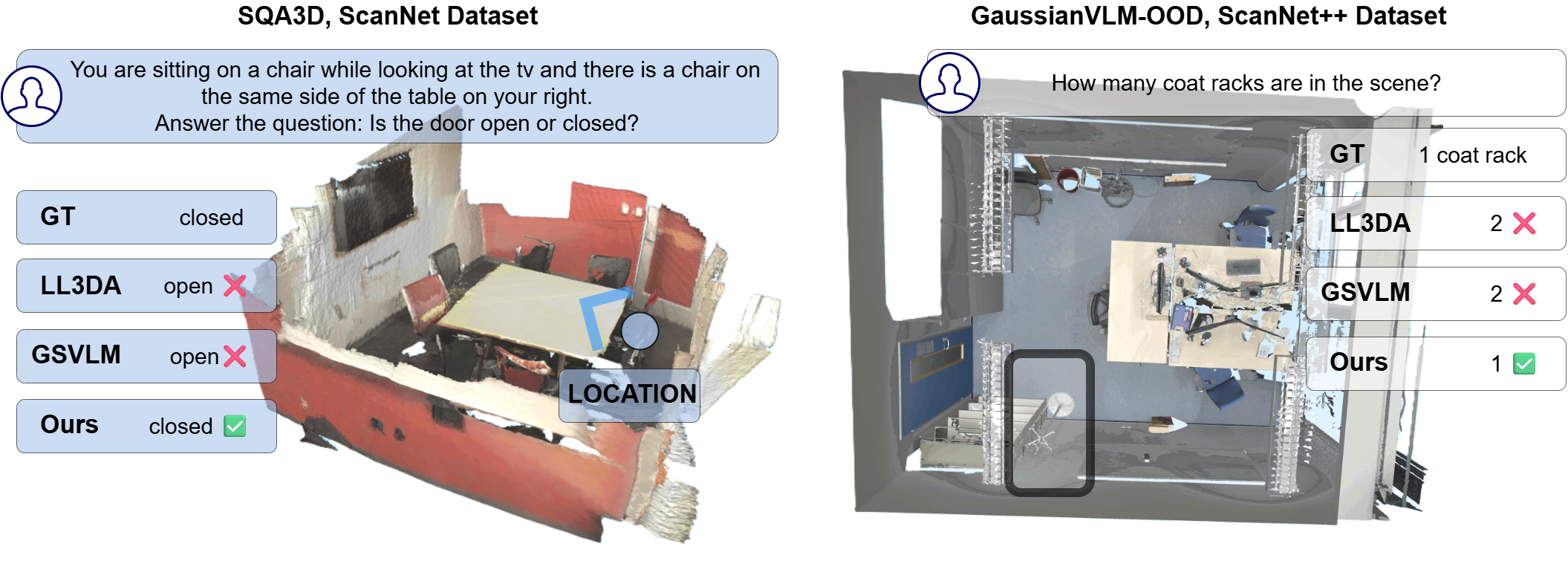}
    \caption{Qualitative results on situated question answering (SQA3D benchmark) and out-of-distribution generalization (GaussianVLM-OOD benchmark). Our model achieves better performance on both benchmarks, showing visual understanding capabilities both on in-domain ScanNet scenes, and on out-of-domain ScanNet++ scenes.}
    \label{fig:quali2}
\end{figure}

\begin{table}[h]
\centering
\caption{Training hyperparameters for P$^3$-VLM.}
\begin{tabular}{lc}
\toprule
\textbf{Hyperparameter} & \textbf{Value} \\
\midrule
\multicolumn{2}{c}{\textit{Optimization}} \\
Optimizer           & AdamW \\
Learning rate       & $3\times10^{-5}$ \\
$(\beta_1, \beta_2)$  & $(0.9,\ 0.999)$ \\
Weight decay        & $0.05$ \\
LR schedule         & Linear warmup + cosine decay \\
Warmup steps        & $200$ \\
Gradient clip (norm)& $5.0$ \\
\midrule
\multicolumn{2}{c}{\textit{Batching}} \\
Batch size (per GPU)        & $16$ \\
Gradient accumulation steps & $4$ \\
Effective batch size        & $64$ \\
\midrule
\multicolumn{2}{c}{\textit{Language model}} \\
Backbone            & Qwen3-VL-4B-Instruct \\
Max input length    & $2048$ \\
Max output length   & $512$ \\
\midrule
\multicolumn{2}{c}{\textit{LoRA}} \\
Rank ($r$)          & $16$ \\
Alpha ($\alpha$)    & $32$ \\
Dropout             & $0.05$ \\
Target modules      & \texttt{q, k, v, o\_proj} \\
\bottomrule
\end{tabular}
\label{tab:hyperparams_supp}
\end{table}

\section{Further Ablations}
\label{supp_sec:ablations2}
Complementary to the tri-pathway design ablation provided in Sec. \ref{sec:ablations}, we also evaluate specific design decisions within the location pathway and sparsifier components of P$^{3}$-VLM. 

\paragraph{Location Pathway Design.}
For the location pathway we consider only the location-based captioning task, as the other tasks do not utilize the pathway. First, we ablate whether scene-context information improves local information extraction. The results demonstrate better location understanding on raw voxel features ($F$) compared to utilization of already sparsified and refined features ($\mathbf{S}$).We attribute this to the sparsifier smoothing voxel features through spatial self-attention, which reduces their locality -- precisely the property the location pathway relies on. 
We also ablate the extraction strategy comparing our MLP-based approach to FiLM location conditioning, with exact details on the FiLM alternative provided in the Supplementary. Results show that processing raw voxel features ($F$) with an MLP strategy delivers best results. 

\paragraph{Sparsification of Visual Tokens.}
We also ablate the role of positional encodings (PE) for increasing scene-level awareness within the sparsification process. We evaluate the sparsifier module with 3D RoPE PE, learned absolute PE added to voxel features, and no PE. 3D RoPE yields the strongest performance, confirming that explicit spatial structure in the global features is critical.

\begin{table}[t]
    \centering
    \caption{\textbf{Location pathway and Sparsifier design ablation studies.} We ablate the point-based variant, P$^3$-VLM (p). \textit{Loc.\ Design}:
    raw vs.\ processed features and FiLM vs.\ MLP extraction.
    \textit{Pos.\ Enc.}: positional encoding strategy in visual sparsification step.}
    \small
    \begin{tabular}{ l | c c | c c c | c c c c c }
    \toprule
        & \multicolumn{2}{c|}{Nr3D}
        & \multicolumn{3}{c|}{ScanQA}
        & \multicolumn{5}{c}{SQA3D} \\
        \cmidrule{2-3} \cmidrule{4-6} \cmidrule{7-11}
        & Acc & Sim & EM-R & M & R & EM-R & B4 & C & M& R \\
    
    \midrule
    \multicolumn{11}{c}{\textbf{Location Sparsifier Design}}\\
    \midrule
     Processed + MLP    &42.3 & 47.1& -  &  -  &  -  &  -  &  -  &  -  &  -  &  -  \\
     Raw + FiLM      & 48.5& 46.4& -  &  -  &  -  &  -  &  -  &  -  &  -  &  - \\
    \midrule
    \multicolumn{11}{c}{\textbf{Sparsifier Design: Positional Encoding}}\\
    \midrule
    None                  & 47.9 & 43.7& 29.3& 18.8 & 31.1 & 46.6 &  28.7 & 114.5 & 23.4 & 44.4 \\
    Learned absolute      & 49.4 & 46.0 &26.9 & 20.9 & 32.1 & 45.3 & 21.6 & 113.8 & 23.3 & 43.8 \\
    \midrule
    \textbf{Full model}   & 52.1 &46.0 & 31.9 & 22.9& 34.8 & 51.8 &30.58 &130.4 &26.59 &50.32\\
    \bottomrule
    \end{tabular}
    \label{tab:ablations2}
\end{table}

\section{P$^3$-VLM Architecture Details}
\label{supp_sec:architecture}

\subsection{Location Pathway Designs.} We investigated two different location pathway designs -- FiLM- and MLP-based, Both first identify the nearest voxel to the target location, retrieving its feature
vector as a local scene descriptor. The first strategy, \textbf{FiLM-based extraction}, conditions this voxel feature on the queried location via Feature-wise Linear
Modulation~\cite{perezfilm}: we use a shallow location encoder MLP (2 linear layers) to map the 3D point coordinates $\mathbf{p}$ to
affine parameters $(\gamma, \beta) \in \mathbb{R}^D \times \mathbb{R}^D$. Those are afterwards used to modulate the retrieved voxel feature $\mathbf{F}_i \in \mathbb{R}^D$ as
$\mathbf{T}_\text{loc} = \sigma(\gamma) \odot \mathbf{F}_i + \beta$,
where $\sigma(\cdot)$ denotes the sigmoid function, applied to $\gamma$ to act as
a soft selection mask over feature dimensions.  We note that all notation is introduced in Sec. \ref{sec:method}.

We note that the MLP-based variant is introduced in Sec. \ref{sec:method}. The MLP structure we adopt is four-layer MLP with GELU activations.

\subsection{Sparsifier RoPE Details}
The spatial refinement is handled by 4 self-attention layers where both queries and keys are rotated by 3D
Rotary Position Embeddings.
For head dimension $d_{h}$ (divisible by 6), per-axis frequencies
are:
\begin{equation}
    \theta_{i,\,\text{axis}}
        = \frac{\mathrm{pos}_\text{axis}}{1000}
          \cdot \mathrm{base}^{-2i/d_{h}},
    \quad i = 0,\ldots,\tfrac{d_{h}}{6}-1,
\end{equation}
with $\mathrm{base}=10{,}000$.

\subsection{Comparison to the GaussianVLM Architecture}
\label{supp_sec:vs_gaussianvlm}
P$^3$-VLM and GaussianVLM~\cite{halacheva2025gaussianvlm} are both scene-centric 3D GS VLMs, but are built around different priorities, from which their architectural differences follow.
GaussianVLM is \emph{semantics-oriented}: it fuses language into the 3D backbone itself (per-Gaussian SigLIP2 features), so every downstream token is semantically grounded before sparsification. This favors scene-level tasks and semantic VQA, where broad semantic understanding matters more than precise spatial extent.
P$^3$-VLM is \emph{geometry-oriented}, designed to strengthen grounding, which leads to the following design decisions:
\begin{itemize}
    \item \textbf{Context pathway.} A dedicated, unconditional pathway explicitly recovers global scene semantics; GaussianVLM has no counterpart.
    \item \textbf{Task pathway.} In P$^3$-VLM, voxel tokens query the text, and the result is applied as a residual correction to the visual features \emph{before} compression, i.e., language reshapes geometry. In GaussianVLM, text tokens query and directly select visual tokens, i.e., language selects semantics.
    \item \textbf{Location pathway.} P$^3$-VLM derives the local token with an MLP over the concatenated geometric and semantic features at the queried point, explicitly learning location understanding jointly with content. GaussianVLM retrieves the per-Gaussian semantic feature nearest to the query point, without a comparable geometric grounding mechanism.
    \item \textbf{Positional information.} P$^3$-VLM fuses positional information early via 3D RoPE, reinforcing geometry throughout the visual processing; GaussianVLM adds position as a final Fourier embedding, secondary to semantics.
\end{itemize}

\subsection{Point Cloud Encoder Variant}
\label{supp_sec:pc_variant}
To assess whether P$^3$-VLM generalizes beyond 3D GS inputs, we replace the SceneSplat encoder with the self-supervised point cloud encoder Sonata~\cite{wu2025sonata} (P$^3$-VLM-PC), keeping the tri-pathway design and training protocol unchanged. Results are shown in Tab.~\ref{tab:pc_variant}. Owing to Sonata's large-scale pre-training, P$^3$-VLM-PC achieves strong location-grounded captioning results under point prompts (Nr3D: $58.1\%$, ScanRefer: $63.3\%$ Acc), outperforming the 3D GS variant. However, Sonata features are not language-aligned, unlike SceneSplat features, which results in lower performance on the VQA benchmarks. This shows that the tri-pathway design transfers to other 3D representations, with the choice of encoder trading off grounding against language-aligned scene understanding.

\begin{table}[h]
    \centering
    \caption{\textbf{P$^3$-VLM (p) with a 3D GS (SceneSplat) vs.\ point cloud (Sonata) encoder} on grounded captioning (Nr3D, ScanRefer) and VQA (ScanQA, SQA3D).}
    \small
    \setlength{\tabcolsep}{4pt}
    \begin{tabular}{l | c c c c | c c c | c c c c c}
    \toprule
      & \multicolumn{2}{c}{Nr3D} & \multicolumn{2}{c|}{ScanRefer} & \multicolumn{3}{c|}{ScanQA} & \multicolumn{5}{c}{SQA3D}\\
      \cmidrule{2-3} \cmidrule{4-5} \cmidrule{6-8} \cmidrule{9-13}
      Encoder & Acc & Sim & Acc & Sim & EM-R & M & R & EM-R & B4 & C & M & R\\
      \midrule
      SceneSplat (3D GS) & 52.1 & 46.0 & 56.1 & 54.2 & \textbf{31.9} & 22.6 & \textbf{35.3} & \textbf{51.8} & \textbf{30.6} & \textbf{130.4} & \textbf{26.6} & \textbf{50.3}\\
      Sonata (PC) & \textbf{58.1} & \textbf{47.3} & \textbf{63.3} & \textbf{54.8} & 30.5 & \textbf{22.8} & 33.9 & 48.3 & 16.8 & 126.9 & 23.8 & 48.8\\
      \bottomrule
    \end{tabular}
    \label{tab:pc_variant}
\end{table}

\section{Datasets Information}

\paragraph{Point Cloud and 3D GS Representations.}
Our method uses 3D GS scene input, differing from the point cloud input consumed by other methods. To ensure maximal similarity in setup to the point cloud-based methods, we use 3D GS representations with Gaussian centers initialized at the point locations from the original point clouds ("Fixed XYZ" strategy). This guarantees the exact same geometrical properties under the two setups. Our 3D GS data is consistent to the one utilized within the GaussianVLM experiments and is obtained from the GaussianWorld dataset \cite{li2025scenesplat}.

\paragraph{Dataset Licenses and Terms of Use.}
We report the licenses and terms of use for all datasets in Table~\ref{tab:dataset-licenses} and adhere strictly to each. All datasets are publicly released. We draw attention to ScanNet~\cite{dai2017scannet} and ScanNet++~\cite{yeshwanth2023scannet++} being subject to custom terms of use, requiring separate inspection. All other datasets are released under standard open-source licenses. The 3D GS representations of ScanNet and ScanNet++ scans are obtained from the respective subsets under GaussianWorld (SceneSplat-7K) and inherit their respective terms.

\begin{table}[h]
\centering
\caption{Licenses and terms of use for the used datasets.}
\small
\begin{tabular}{l|l}
\toprule
\textbf{Dataset} & \textbf{License / Terms of Use} \\
\midrule
ScanNet \cite{dai2017scannet} & ScanNet Terms of Use\footnotemark[1] \\
ScanRefer \cite{chen2020scanrefer} & Creative Commons BY-NC-SA 3.0 \\
ScanQA \cite{scanqa_22_cvpr} & Apache 2.0 \\
SQA3D \cite{ma2022sqa3d}& Apache 2.0 \\
Nr3D \cite{achlioptas2020referit_3d}& MIT \\
ScanNet++ \cite{yeshwanth2023scannet++} & ScanNet++ Terms of Use\footnotemark[2] \\
GaussianWorld (SceneSplat-7K) \cite{li2025scenesplat}& Same as ScanNet, 
ScanNet++ \\
\bottomrule
\end{tabular}
\label{tab:dataset-licenses}
\end{table}

\footnotetext[1]{Available at \url{https://kaldir.vc.in.tum.de/scannet/ScanNet_TOS.pdf} (last accessed: 03/05/2026).}
\footnotetext[2]{Available at \url{https://kaldir.vc.in.tum.de/scannetpp/static/scannetpp-terms-of-use.pdf} (last accessed: 03/05/2026).}

\section{Broader Impact}
Our work aims to make a step towards VLMs with strong visual understanding capabilities, showing strong robustness and out-of-distribution generalization. With this we invision real-world utilization of 3D VLMs for applications that require free-form, open-world, spatially grounded reasoning, such as robotics, assistive technologies, and AR/VR systems. 

We also note the potential environmental implications of training VLM models such as ours. While P$^3$-VLM is designed with sparsification modules and scene downsampling, its training still requires significant compute resources. With this in mind, we underscore the necessity of research into environmentally-friendly training approaches.
\clearpage



\end{document}